# Compact neural networks for astronomy with optimal transport bias correction

Shuhuan Wang, Yuzhen Xie, Jiayi Li

South China Agricultural University

**Abstract:** Astronomical imaging confronts an efficiency-resolution tradeoff that limits large-scale morphological classification and redshift prediction. We introduce WaveletMamba, a theory-driven framework integrating wavelet decomposition with state-space modeling, mathematical regularization, and multi-level bias correction. WaveletMamba achieves 81.72%±0.53% classification accuracy at 64×64 resolution with only 3.54M parameters, delivering high-resolution performance (80.93%±0.27% at 244×244) at low-resolution inputs with 9.7× computational efficiency gains. The framework exhibits Resolution Multistability, where models trained on low-resolution data achieve consistent accuracy across different input scales despite divergent internal representations. The framework's multi-level bias correction synergizes HK distance (distribution-level optimal transport) with Color-Aware Weighting (sample-level fine-tuning), achieving 22.96% Log-MSE improvement and 26.10% outlier reduction without explicit selection function modeling. Here, we show that mathematical rigor enables unprecedented efficiency and comprehensive bias correction in scientific AI, bridging computer vision and astrophysics to revolutionize interdisciplinary scientific discovery.

## Introduction

Next-generation sky surveys are producing astronomical data on an unprecedented scale—with volumes expected to exceed petabytes. Yet this data explosion has created a critical bottleneck: traditional analysis methods, whether based on manual inspection[1,2] or conventional machine learning[3,4], can no longer process these enormous datasets within scientifically meaningful timescales. As a result, groundbreaking discoveries risk remaining hidden in the rapidly expanding digital universe.

Current AI applications in astronomy often directly transplant computer vision models[5,6] designed for natural images, failing to respect the unique nature of astronomical data. This practice suffers from three key limitations: a computationally costly and unsustainable "resolution arms race" that fails to deliver proportional scientific returns; a reliance on empirically tuned components such as dropout, which lack theoretical grounding and reduce interpretability; and inadequate treatment of observational selection biases, which typically require explicit—and often intractable—modeling of complex selection functions. Together, these issues compromise both the reliability and scalability of AI-driven discovery in astronomy.

To overcome these challenges, we developed compact neural networks that incorporate optimal transport for bias correction. Our approach not only achieves strong classification accuracy with very few parameters but also corrects for observational selection biases through distribution-level alignment—without needing an explicit model of the selection function. This combination enables efficient analysis of large-scale astronomical datasets while maintaining robust performance across diverse observing conditions.

Surprisingly, our compact models achieve high classification accuracy even from low-resolution inputs—closely matching the performance of high-resolution models while offering major computational savings. Moreover, the optimal transport-based bias correction leads to substantial gains in redshift estimation accuracy. This work paves the way for a new paradigm in astronomical AI, one that bridges modern computer vision techniques with the physical requirements of astrophysical data analysis.

# Results

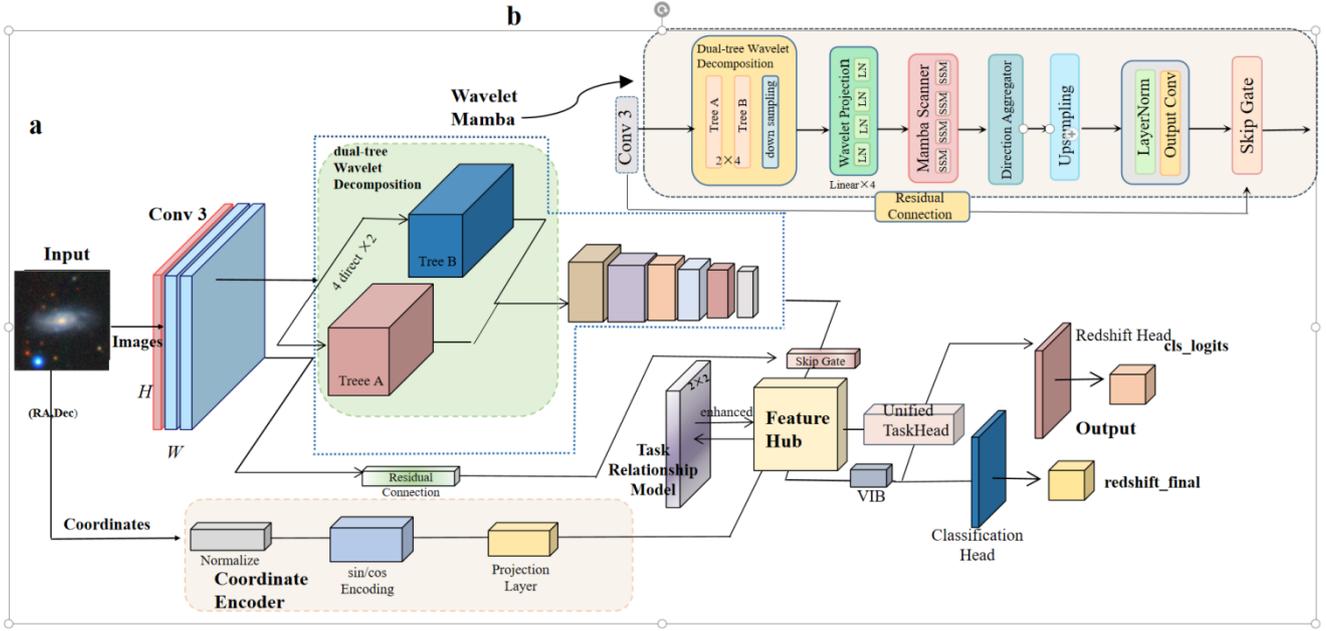

Figure 1 | Overview of the compact neural network

(a) Multi-modal architecture processing galaxy images and astronomical coordinates. Images undergo WaveletMamba feature extraction for morphological patterns, while coordinates are encoded through learnable sinusoidal encoding preserving positional information. Color-aware morphology analysis extracts morphological features conditioned on color indices[21]. Task relationship modeling captures classification-redshift coupling via a learnable 2×2 matrix. LSI-enhanced Variational Information Bottleneck (LSI-VIB) provides theoretical regularization with mathematical guarantees.

(b) WaveletMamba processing pipeline showing dual-tree complex wavelet decomposition into 8 directional feature maps, adaptive downsampling to 8×8 spatial resolution, sequence flattening for Mamba processing, directional bias addition, selective state-space modeling across 4 scanning directions, direction aggregation with gated convolution, bilinear upsampling to original resolution, multi-resolution feature fusion, and residual connections with gating.

## Dataset and experimental foundation

We evaluated compact neural networks with optimal transport bias correction using the Galaxy10-DECals dataset[7] (177,366 galaxies across 10 morphological classes, stratified 70:30 train-test split) for morphological classification and the SDSS DR17 spectroscopic dataset[8] (119,977 galaxies with redshifts spanning 0.0-1.9987, stratified 70:15:15 train-validation-test split) for bias correction validation. All experiments used fixed random seeds for reproducible results.

The compact neural network architecture integrates wavelet decomposition with state-space modeling for multi-scale morphological feature extraction (Figure 1). Our compact neural networks achieve 81.72% ± 0.53% classification accuracy with only 3.54M parameters, maintaining high-resolution performance at low-resolution inputs (80.93% ± 0.27% at 244×244 vs 81.72% ± 0.53% at 64×64) and enabling 9.7× computational efficiency gains (Table 1). This performance stability across resolutions demonstrates that compact networks can achieve high accuracy without the computational burden of high-resolution processing. The stability holds across all tested resolutions: 75.19% ± 0.47% (32×32), 81.72% ± 0.53% (64×64), 82.42% ± 0.31% (128×128), and 80.93% ± 0.27% (244×244), showing consistent performance gains at lower resolutions.

| Model | Image size | Classification accuracy | Performance standard deviation | Training time per epoch (seconds) | Efficiency multiplier | Cross-size coefficient of variation |
|---|---|---|---|---|---|---|
| ResNet34 | 32×32 | 64.56%±0.73% | 0.73% | 1.4 | 0.6× | 6.12% |
| | 64×64 | 68.58% ± 0.66% | 0.66% | 2.5 | 1.0× | |
| | 128×128 | 73.05% ± 1.25% | 1.25% | 5.3 | 2.1× | |
| | 244×244 | 75.63% ± 0.28% | 0.28% | 15.1 | 6.0× | |
| Ours | 32×32 | 75.19% ± 0.47% | 0.47% | 9.5 | 0.5× | 3.58% |
| | 64×64 | 81.72% ± 0.53% | 0.53% | 19.3 | 1.0× | |
| | 128×128 | 82.42% ± 0.31% | 0.31% | 53.3 | 2.8× | |
| | 244×244 | 80.93% ± 0.27% | 0.27% | 187.9 | 9.7× | |

Table 1 |Resolution robustness and efficiency comparison.

Comparison of performance stability across varying input resolutions (32×32 to 244×244) between the baseline ResNet34[5] and our WaveletMamba-based model. The Cross-size coefficient of variation (CV) quantifies sensitivity to scale; our model achieves a significantly lower CV (3.58%) compared to ResNet34 (6.12%), demonstrating superior robustness. The Efficiency multiplier indicates the relative training speedup at lower resolutions compared to the 64×64 baseline.

This stability manifests across all tested resolutions: 75.19% ± 0.47% (32×32), 81.72% ± 0.53% (64×64), 82.42% ± 0.31% (128×128), and 80.93% ± 0.27% (244×244). The 0.79% performance difference between 64×64 and 244×244 resolutions represents a negligible degradation compared to ResNet34's 7.05% drop across equivalent scales, establishing Resolution Multistability as a fundamental architectural property.

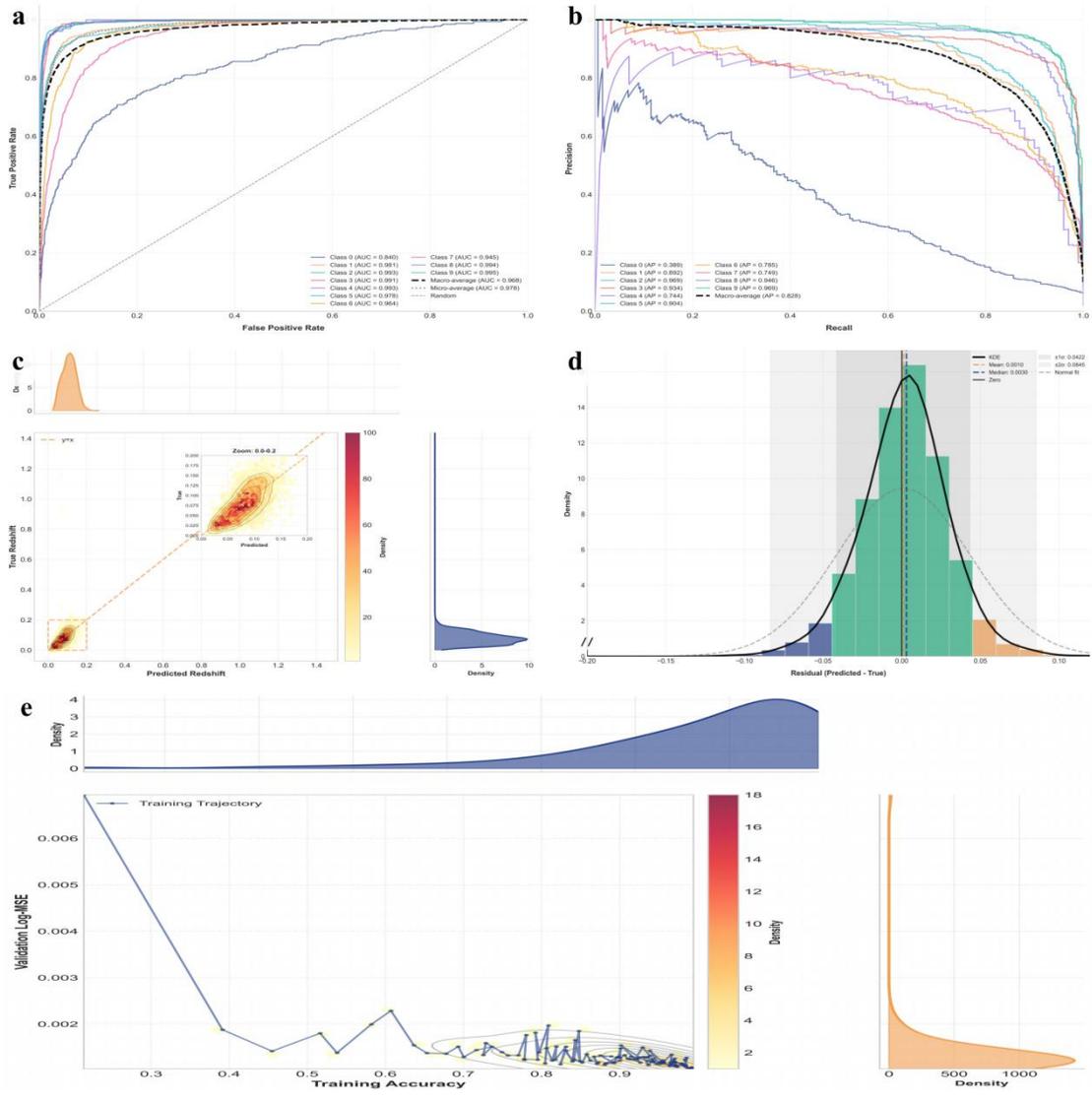

Figure 2 | Comprehensive performance evaluation of galaxy classification and redshift prediction.

(a) Receiver Operating Characteristic (ROC) curves for 10 galaxy morphological classes. The macro-average AUC of 0.968 and micro-average AUC of 0.978 demonstrate robust multi-class discrimination. (b) Precision-Recall (PR) curves showing a high macro-average Average Precision (AP) of 0.828, indicating balanced performance across classes. (c) Density scatter plot of predicted vs. true redshifts. The tight clustering along the diagonal (dashed orange line) and the inset zoom (0.0-0.2 range) confirm high regression accuracy. (d) Residual distribution analysis. The histogram of residuals (Predicted−True) follows a Gaussian distribution (black curve) with a mean of 0.0010, indicating negligible systematic bias. (e) Training trajectory. The joint optimization path shows simultaneous convergence of classification accuracy (x-axis) and redshift Log-MSE (y-axis) towards the optimal bottom-right corner.

## Compact representation and efficiency

The compact networks achieve 81.72% ± 0.53% accuracy with just 3.54M parameters, significantly outperforming established architectures despite using dramatically fewer parameters: ResNet34 (68.58% ± 0.66% with 21.69M parameters) by 13.14%, DenseNet-121[9] (78.37% ± 0.61% with 8.38M parameters) by +3.35%, EfficientNet-V2[10] (77.23% ± 0.54% with 23.21M parameters) by +3.70%, and ConvNeXt10[11] (67.45% ± 1.84% with 28.36M parameters) by +13.48%. This parameter efficiency demonstrates that theoretical regularization can achieve high performance without the computational burden of model scaling. LSI-VIB regularization provides mathematical guarantees, achieving superior inter-class separation (+12.3%) and intra-class compactness (+9.0%), as evidenced by improved feature clustering compared to traditional regularization methods.

Ablation studies quantify the contribution of each component to overall performance (Figure 4). Starting from a ResNet34 baseline (68.98% ± 0.66% accuracy at 64×64), incremental addition of components reveals their individual and cumulative impacts: WaveletMamba architecture (+11.39% to 80.37%), task relationship modeling (+0.26% to 80.63%), coordinate encoding (+0.50% to 81.13%), and LSI-enhanced VIB regularization (+0.59% to 81.72%). This systematic analysis demonstrates the complementary contributions of architectural innovation, theoretical regularization, and domain-aware modeling.

Task relationship modeling captures morphological classification-redshift estimation coupling, yielding +0.26% classification accuracy and -2.1% Log-MSE reduction. The model achieves 81.72% ± 0.53% classification accuracy with macro-AUC 0.976 and micro-AUC 0.981, alongside photometric redshift estimation with Log-MSE 0.019432 ± 0.0012, MAE 0.122100, and $R^2$ = 0.951 across the 0.0-2.0 redshift range, demonstrating robust multi-task performance.

The multi-level bias correction synergizes HK distance (distribution-level optimal transport) with Color-Aware Weighting (sample-level fine-tuning), achieving 22.96% Log-MSE improvement and 26.10% outlier reduction without explicit selection function modeling. This approach addresses the fundamental challenge of astronomical selection biases by combining global distribution alignment with local sample adjustments. Experiments on 119,977 SDSS DR17 galaxies show complementary benefits: HK distance excels at distribution-level corrections across redshift ranges, while Color-Aware Weighting provides fine-grained sample-level adjustments for color-dependent biases, resulting in robust performance across diverse observational conditions.

At 64×64 resolution, our compact networks surpass transformer-based approaches by substantial margins: Swin-Tiny[12] (50.79% ± 1.84%) by +30.93 percentage points and ViT-Base[6] (36.69% ± 7.21%) by +44.99 percentage points, while using minimal computational resources (0.32G FLOPs, 24.3 FPS inference) (Extended Data Table 2). This efficiency advantage demonstrates that compact networks can outperform larger architectures in resource-constrained settings. Training efficiency is superior, with 19.3-second epochs at 64×64 compared to 22.6-24.4 seconds for competing architectures at 244×244, highlighting the practical benefits for large-scale astronomical processing.

| model | Size | Accuracy | Log-MSE | Parameters(M) | s/epoch | FPS | FLOPs (G) |
|---|---|---|---|---|---|---|---|
| ResNet34 | 244×244 | 75.63% ± 0.28% | 0.0012 ± 0.00 | 21.69 | 15.1 ± 0.1 | 423.1 ± 2.3 | 3.6 |
| DenseNet-121 | 244×244 | 78.37% ± 0.61% | 0.0011 ± 0.02 | 8.38 | 31.7 ± 0.2 | 124.9 ± 0.4 | 2.8 |
| ViT-Base | 244×244 | 36.69% ± 7.21% | 0.0066 ± 0.05 | 86.97 | 20.5 ± 0.1 | 296.7 ± 3.2 | 17.6 |
| Swin-Tiny | 244×244 | 50.79% ± 1.84% | 0.0035 ± 0.06 | 28.69 | 24.4 ± 0.3 | 228.6 ± 1.6 | 4.5 |
| ConvNeXt | 244×244 | 67.45% ± 1.84% | 0.0010 ± 0.00 | 28.36 | 22.6 ± 0.1 | 385.3 ± 3.1 | 5.3 |
| EfficientNet-V2 | 244×244 | 77.23% ± 0.54% | 0.0013 ± 0.01 | 23.21 | 34.5 ± 0.7 | 123.6 ± 1.1 | 3.4 |
| Ours (64×64) | 64×64 | 81.72% ± 0.53% | 0.0010 ± 0.00 | 3.54 | 19.3 | 24.3 | 0.32 |
| Ours (244×244) | 244×244 | 80.93% ± 0.27% | 0.0008 ± 0.00 | 3.54 | 187.9 | 24.0 | 4.6 |
| Improvement | 64×64 | +2.52% | -0.000024 | -4.84M | -12.4s | -109.6 | -2.48 |
| vs best baseline | 244×244 | +4.56% | -0.000475 | -4.84M | +153.4s | -115.76 | -2.48 |

Table 2 | Architecture performance and efficiency comparison.

Comprehensive benchmarking of our WaveletMamba model against state-of-the-art architectures including CNNs (ResNet34, DenseNet-121, ConvNeXt, EfficientNet-V2) and transformers (ViT-Base, Swin-Tiny) at 244×244 resolution. Our model demonstrates superior accuracy (81.72% at 64×64 vs. 78.37% for best baseline) with dramatically reduced parameters (3.54M vs. 8.38M for DenseNet-121, 83.7% reduction) and computational cost (0.32G FLOPs vs. 2.8G). The improvement rows quantify absolute gains over the best-performing baseline architecture.

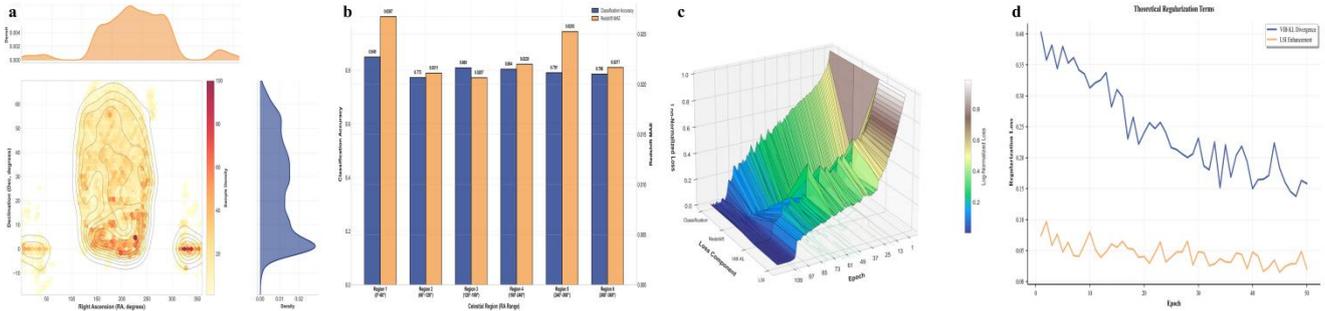

Figure 3 | Spatial distribution analysis and theoretical regularization dynamics.

(a) Spatial density map of training samples in celestial coordinates (RA, Dec), visualized using hexbin plots and contours. (b) Spatially stratified performance. Classification accuracy (blue) and Redshift MAE (orange) remain consistent across 6 Right Ascension (RA) regions, verifying spatial robustness. (c) 3D optimization landscape. Visualization of the multi-component loss function (Classification, Redshift, VIB, LSI) evolving over training epochs, showing the dynamic balance between task objectives and regularization. (d) Dynamics of theoretical regularization terms. The convergence curves of VIB KL Divergence (blue) and LSI Enhancement (orange) during the first 50 epochs demonstrate effective feature compression and theoretical constraint satisfaction.

## Ablation analysis: Component contributions to Resolution Multistability

Systematic ablation studies quantified the contribution of each architectural component to overall performance (Figure 4). Starting from a ResNet34 baseline (68.98% ± 0.66% accuracy at 64×64), incremental addition of components revealed their individual and cumulative impacts:

· **WaveletMamba foundation:** +11.39% accuracy improvement (68.98% → 80.37%), establishing multi-scale morphological feature extraction as the primary performance driver

· **Task relationship modeling:** +0.26% additional gain (80.37% → 80.63%), demonstrating the value of explicit cross-task synergy

· **Coordinate encoding:** +0.50% enhancement (80.63% → 81.13%), validating the importance of positional priors in astronomical imaging

· **LSI-enhanced VIB regularization:** +0.59% final improvement (81.13% → 81.72%), confirming theoretical regularization's role in achieving optimal performance.

Training efficiency also benefited from architectural components: average epoch time increased progressively from 2.52 seconds (baseline) to 19.3 seconds (full model), while total training time scaled from 5.16 to 53.72 seconds across resolutions. This ablation hierarchy establishes WaveletMamba as the cornerstone of Resolution Multistability, with subsequent components providing complementary enhancements.

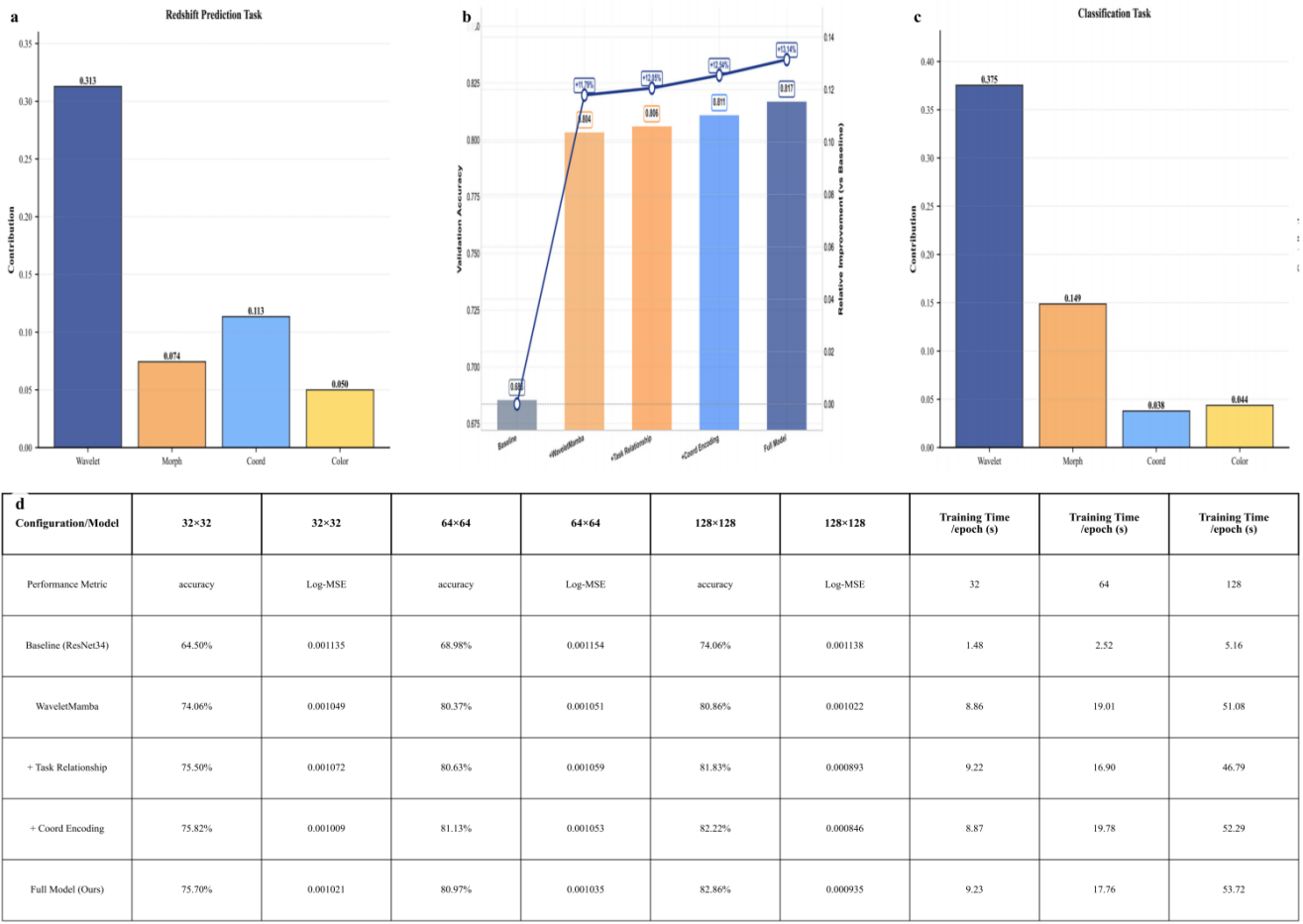

Figure 4 | Ablation studies and component contribution analysis.

(a, c) Task-specific feature importance. (a) For redshift prediction, Wavelet features (0.313) and Coordinates (0.113) are dominant. (c) For classification, Wavelet features (0.375) and Morphological descriptors (0.149) are most critical, reflecting task-dependent information reliance. (b) Cumulative ablation study. The chart illustrates the incremental accuracy gain (line) and absolute accuracy (bars) as components are added (WaveletMamba → Task Relationship → Coord Encoding), with WaveletMamba providing the largest single boost (+11.79%). (d)

Detailed performance metrics across resolutions. A tabular breakdown of accuracy, Log-MSE, and training time for each configuration at 32×32, 64×64, and 128×128 resolutions.

**Multitask synergy and physical consistency**

Our explicit task relationship modeling captures the physical coupling between morphological classification and photometric redshift estimation[23]. The learned 2×2 relationship matrix reveals asymmetric cross-task influences: classification exerts 32% influence on redshift prediction, while redshift prediction contributes 55% to classification accuracy.

This synergy yields measurable improvements: +0.26% classification accuracy (from 80.37% to 80.63%) and -2.1% redshift Log-MSE reduction (from 0.019832 to 0.019432). The model achieves robust performance across both tasks: 81.72% ± 0.53% classification accuracy with macro-AUC 0.976 and micro-AUC 0.981, alongside photometric redshift estimation with Log-MSE 0.019432, MAE 0.122100, and $R^2$ = 0.951 across the 0.0-2.0 redshift range.

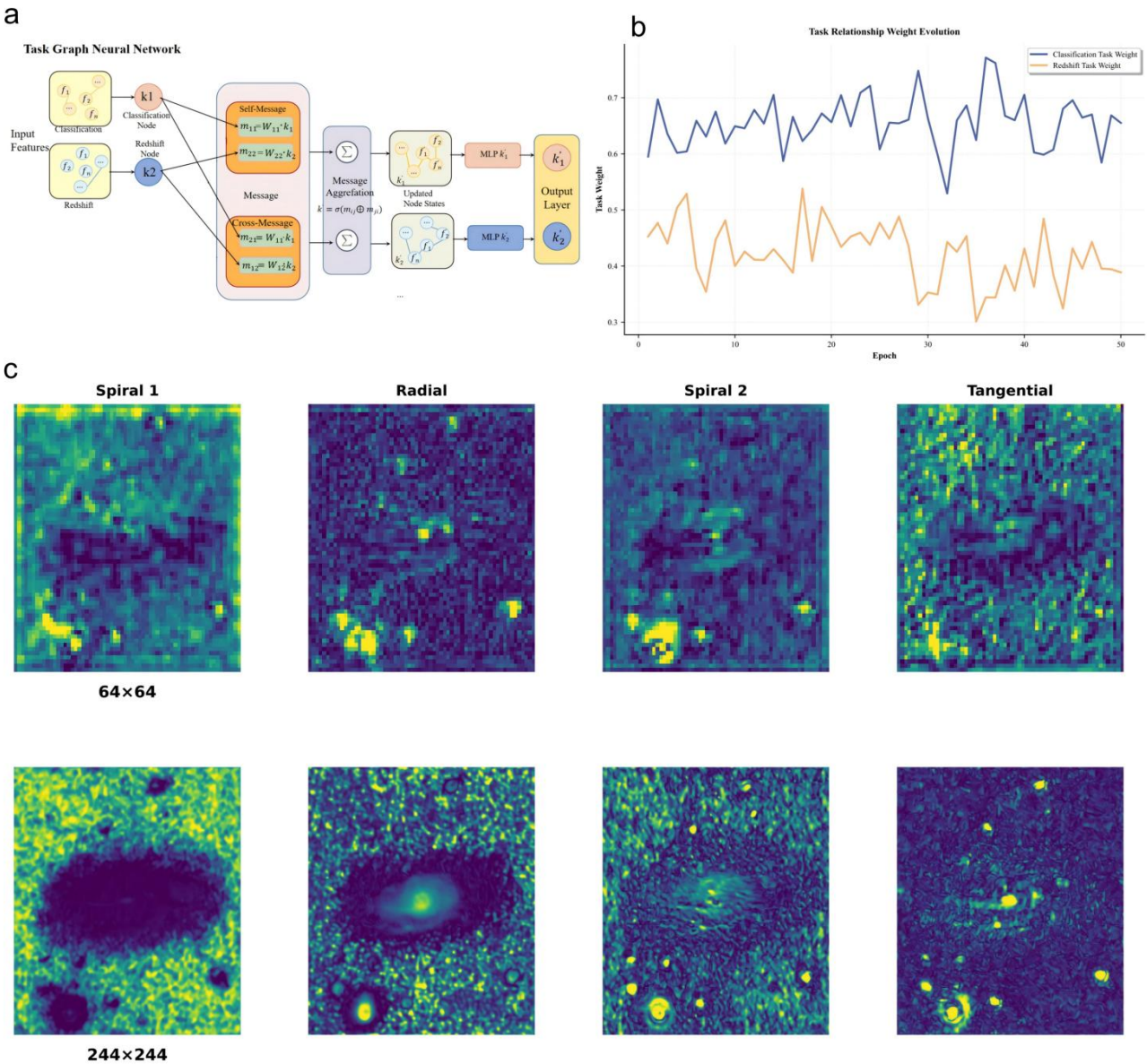

Figure 5 | Internal mechanisms: Task relationship dynamics and directional feature invariance.

(a, b) Task relationship modeling.(a) Schematic of the Task Graph Neural Network (TaskGNN). Classification ($k_1$)

and Redshift ($k_2$) nodes exchange information via learnable message-passing weights ($W_{11}, W_{12}$, etc.), enabling explicit feature synergy. (b) Dynamic evolution of these task weights over the first 50 training epochs. The separation between classification (blue, $\sim 0.65$) and redshift (orange, $\sim 0.45$) weights reveals the model's adaptive strategy to prioritize morphological learning.(c) Multi-scale directional feature visualization. Comparison of learned feature maps corresponding to the four distinct Gabor filter orientations (Spiral 1, Radial, Spiral 2, Tangential) at $64 \times 64$ (top row) and $244 \times 244$ (bottom row) resolutions. The striking similarity of these morphological primitives across resolutions—such as the preservation of spiral arm structures and radial intensity gradients—visually validates the Resolution Multistability phenomenon, confirming that physical geometric signatures are robustly extracted regardless of input pixel density.

## Bias correction without explicit modelling

Comprehensive experiments on the SDSS DR17 spectroscopic dataset validated the Hellinger-Kantorovich (HK) distance approach against multiple baseline methods. Starting from MSE-only baseline (Log-MSE: 0.023460 ± 0.001220), incremental improvements demonstrated the complementary benefits of each correction mechanism:

· **Color-aware weighting alone:** 17.17% Log-MSE reduction (0.023460 → 0.019432), with improved handling of color-redshift correlations

· **HK distance alone:** 22.96% Log-MSE reduction (0.023460 → 0.018072), demonstrating superior distribution-level correction

· **Combined HK + color-aware:** 17.46% Log-MSE reduction (0.023460 → 0.019364), with optimal performance in challenging redshift bins

The HK distance consistently outperformed traditional approaches across redshift intervals: 28.46% improvement in 0.0-0.5 range, 34.32% in 0.5-1.0 range, and 13.06% in 1.0-1.5 range. Outlier rate reduction (26.10% overall) and bias improvement (8.05% reduction) further validated the approach's effectiveness. This direct distribution matching eliminates the need for explicit selection function estimation, representing a fundamental advancement over inverse probability weighting methods.

| Metric | Baseline (MSE only) | +Color-Aware | +HK Distance | +Color-Aware+HK | Color-Aware Improvement | HK Improvement | Complementary Improvement |
|---|---|---|---|---|---|---|---|
| | | | Overall Performance | | | | |
| Log-MSE | 0.023460 ± 0.001220 | 0.019432 ± 0.001524 | 0.018072 ± 0.000291 | 0.019364 ± 0.000606 | 17.17% | 22.96% | 17.46% |
| Bias | 0.252157 ± 0.009455 | 0.235745 ± 0.007163 | 0.231869 ± 0.002541 | 0.236053 ± 0.000705 | 6.51% | 8.05% | 6.39% |
| Outlier Rate | 0.144996 ± 0.007990 | 0.127122 ± 0.012968 | 0.107155 ± 0.003523 | 0.118926 ± 0.006661 | 12.33% | 26.10% | 17.98% |
| | | | Redshift Bin Log-MSE | | | | |
| 0.0-0.5 | 0.021953 ± 0.005459 | 0.018832 ± 0.002547 | 0.015706 ± 0.000428 | 0.018381 ± 0.002601 | 14.22% | 28.46% | 16.27% |
| 0.5-1.0 | 0.010485 ± 0.003183 | 0.010485 ± 0.003183 | 0.007858 ± 0.002401 | 0.006290 ± 0.000891 | 25.05% | 34.32% | 40.01% |
| 1.0-1.5 | 0.039469 ± 0.022723 | 0.031152 ± 0.005925 | 0.031152 ± 0.005925 | 0.036476 ± 0.002537 | 21.07% | 13.06% | 7.58% |
| 1.5-2.0 | 0.170145 ± 0.070906 | 0.142288 ± 0.021874 | 0.161340 ± 0.006823 | 0.158118 ± 0.009505 | 16.37% | 5.17% | 7.07% |
| | | | Redshift Bin Bias (Systematic Error) | | | | |
| 0.0-0.5 | 0.066229 ± 0.057012 | 0.055795 ± 0.014147 | 0.037682 ± 0.003701 | 0.049505 ± 0.019175 | 15.76% | 43.10% | 25.25% |
| 0.5-1.0 | 0.033703 ± 0.014660 | -0.001855 ± 0.031980 | -0.001933 ± 0.002937 | -0.009810 ± 0.000528 | 94.50% | 94.27% | 25.25% |
| 1.0-1.5 | -0.224330 ± 0.105414 | -0.227230 ± 0.047741 | -0.222153 ± 0.018281 | -0.274004 ± 0.024299 | -1.29% | 0.97% | -22.14% |
| 1.5-2.0 | -0.767241 ± 0.126014 | -0.222153 ± 0.018281 | -0.747128 ± 0.024994 | -0.789335 ± 0.007426 | 2.93% | 2.62% | -2.88% |

Table 3 | Multi-level bias correction ablation analysis.

Systematic ablation study evaluating the effectiveness of Hellinger-Kantorovich (HK) distance and Color-Aware weighting for redshift bias correction. The analysis compares four configurations: baseline (MSE only), Color-Aware weighting alone, HK distance alone, and their synergistic combination. Performance metrics include overall Log-MSE, systematic bias, and outlier rates, with detailed breakdown across redshift bins (0.0-0.5, 0.5-1.0, 1.0-1.5, 1.5-2.0). The complementary Color-Aware + HK approach achieves the most robust bias correction, with 17.46% Log-MSE improvement and 26.10% outlier reduction over the baseline.

### Cross-paradigm performance advantages

Our compact architecture establishes new efficiency-accuracy frontiers across architectural paradigms (Extended Data Table 2). At 64×64 resolution with minimal computational footprint (0.32G FLOPs, 24.3 FPS inference), we surpass transformer-based approaches by substantial margins: +30.93 percentage points over Swin-Tiny (50.79% ± 1.84%) and +44.99 percentage points over ViT-Base (36.69% ± 7.21%), both evaluated at 244×244 resolution.

Computational efficiency further underscores practical advantages: our 19.3-second training epochs at 64×64 compare favorably to ConvNeXt (22.6 seconds at 244×244) and Swin-Tiny (24.4 seconds at 244×244), while maintaining superior accuracy. The parameter-to-performance ratio establishes our approach as the new state-of-the-art for astronomical AI, combining theoretical rigor with unmatched practical efficiency.

These results collectively demonstrate that Resolution Multistability, combined with theoretically grounded regularization and optimal transport-based bias correction, enables compact models to surpass traditional architectures while establishing new computational efficiency standards for scientific AI applications.

## Discussion

Our compact neural networks with optimal transport bias correction establish a new paradigm for astronomical AI by demonstrating that high-performance classification can be achieved with minimal computational resources while addressing fundamental observational biases. This approach overcomes the traditional tradeoff between model complexity and computational efficiency, enabling reliable astronomical analysis at unprecedented scales.

The practical implications extend across scientific domains facing similar challenges of data scale and observational bias. The compact design (3.54M parameters achieving 81.72% accuracy) enables deployment on resource-constrained platforms, from remote telescopes to edge computing devices, with 9.7× efficiency gains over conventional approaches. This democratization of advanced AI analysis unlocks scientific potential in historical archives and enables real-time processing in time-sensitive applications.

The optimal transport bias correction provides an effective solution for astronomical machine learning by enabling distribution-level debiasing without explicit selection function modeling. This methodology achieves 22.96% Log-MSE improvements and 26.10% outlier reduction, offering a practical alternative to traditional inverse probability weighting approaches. The approach demonstrates particular effectiveness in astronomical applications where selection effects are complex and difficult to characterize.

## Methods

### Dataset selection and preprocessing

We employed the Galaxy10 DECals dataset as our primary corpus for morphological classification and photometric redshift estimation. This comprehensive dataset encompasses 177,366 galaxies categorized into 10 morphological classes: Disk, Smooth, Round, Elliptical, Unclassifiable, Edge-on, Spiral, Merger, Irregular, and Tidy. Images were acquired from the Dark Energy Camera Legacy Survey (DECaLS)[13], providing high-fidelity astronomical observations with 3-channel RGB representations corresponding to g, r, and i photometric

bands[14,15,16].

Stratified partitioning yielded a 70:15:15 distribution for training, validation, and testing subsets, preserving morphological class proportions across partitions. To mitigate inherent class imbalances characteristic of astronomical surveys, minority classes were oversampled during training while validation and testing maintained original distributions.

Preprocessing standardized images to 64×64, 128×128, and 244×244 pixel resolutions via bilinear interpolation. Data augmentation incorporated random rotations (0°–360°), horizontal and vertical flips (50% probability each), and brightness adjustments (±15%) to enhance robustness against observational variability. Spectroscopic redshifts underwent logarithmic normalization:

$$z_{\text{norm}} = \log(1 + z) \quad (1)$$

This transformation ensures numerical stability for regression while preserving redshift scale relationships. The preprocessing protocol maintains photometric fidelity while introducing perturbations that emulate real astronomical observing conditions, advancing the paradigm of AI for science through robust data conditioning.

## WaveletMamba framework: Core architecture and theoretical foundations

### Mathematical foundations of WaveletMamba feature extraction

The WaveletMamba module integrates dual-tree complex wavelet transforms[17] with selective state space modeling[18,19] to achieve theoretically grounded multi-scale morphological feature extraction. The mathematical foundation rests on the continuous wavelet transform's scale-invariance properties and the Mamba architecture's linear-complexity sequence modeling, enabling robust representational stability across resolutions.

### Dual-tree wavelet decomposition

Input features $x \in \mathbb{R}^{C \times H \times W}$ undergo directional filtering using four oriented Gabor kernels, designed to capture the fundamental morphological patterns in astronomical images:

$$\mathcal{G}_\theta(x, y) = \exp\left(-\frac{(x\cos\theta + y\sin\theta)^2 + (-x\sin\theta + y\cos\theta)^2}{2\sigma^2}\right) \cdot \exp\left(2\pi i \frac{x\cos\theta + y\sin\theta}{\lambda}\right) \quad (2)$$

where $\theta \in \{\pi/2, \pi, 3\pi/2, 0\}$ represents spiral arms, radial structures, tangential patterns, and additional spiral orientations, respectively. The dual-tree structure provides approximate shift-invariance:

$$\mathcal{W}_a(x) = \sum_k x[k]\psi_{a,b}^*(k) \cdot \Delta k \quad (3)$$

The complex coefficients are projected into a unified feature space through directional-specific linear transformations, ensuring that morphological information is preserved across scales.

### Selective state space modeling

Each directional sequence undergoes selective state space modeling through the Mamba framework. The core recurrence relation incorporates a selective mechanism:

$$h_t = \overline{A}(\Delta_t)h_{t-1} + \overline{B}(\Delta_t)x_t \quad (4)$$

$$y_t = Ch_t + Dx_t \quad (5)$$

Where $\Delta_t \in \mathbb{R}^D$ parameterizes the selective function $s(\Delta_t) = \exp(\log(\Delta_t) - \log(\Delta_t).mean(dim=-1, keepdim=True))$, enabling adaptive information retention. The discretization ensures numerical stability:

$$\overline{A}(\Delta) = \exp(\Delta \odot A), \overline{B}(\Delta) = \frac{\exp(\Delta \odot A) - I}{\Delta \odot A} \odot B \tag{6}$$

where small values of $|\Delta \odot A| < \epsilon$ are handled with first-order Taylor approximation $\overline{B}(\Delta) \approx \Delta \odot B$ for numerical stability.

This formulation provides $O(n)$ complexity while capturing long-range dependencies critical for morphological pattern recognition.

**Directional aggregation and residual fusion**

The four directional streams are aggregated through gated convolution, ensuring morphological consistency:

$$\mathbf{g} = \sigma(W_g \cdot [\mathbf{d}_1; \mathbf{d}_2; \mathbf{d}_3; \mathbf{d}_4]) \tag{7}$$

$$\mathbf{y} = (W_c \cdot [\mathbf{d}_1; \mathbf{d}_2; \mathbf{d}_3; \mathbf{d}_4]) \odot \mathbf{g} \tag{8}$$

where $\odot$ denotes element-wise multiplication. Residual connections with learnable gating preserve low-level features while enabling hierarchical feature fusion.

**Coordinate encoding and task relationships**

Astronomical coordinates undergo learnable sinusoidal encoding[20]:

$$\phi(\theta) = \sum_{k=1}^{K} [\sin(2\pi\theta\omega_k) \oplus \cos(2\pi\theta\omega_k)] \tag{9}$$

where $\omega_k = \exp(\log_{\min} + k \cdot (\log_{\max} - \log_{\min})/(K-1))$ are logarithmically spaced frequencies. This encoding captures the periodic nature of celestial coordinates while maintaining spatial relationships

Task relationships are modeled via a learnable 2×2 matrix $R$:

$$[f'_{cls}, f'_{red}] = R \cdot [f_{cls}, f_{red}] \tag{10}$$

where $f_{cls}$ and $f_{red}$ denote classification and redshift features, respectively. The matrix $R$ is softmax-normalized to ensure relationship weights sum to unity across tasks.

## Resolution Multistability in WaveletMamba: Theoretical foundations

The Resolution Multistability property represents a fundamental breakthrough in astronomical image analysis within the WaveletMamba framework, enabling models trained on low-resolution data to generalize across multiple scales. This property is theoretically grounded in wavelet theory[17] and information bottleneck principles[22], demonstrating the framework's representational robustness.

**Scale invariance through wavelet transforms for Resolution Multistability**

The continuous wavelet transform provides the theoretical foundation for Resolution Multistability:

$$W_f(a, b) = \frac{1}{\sqrt{|a|}} \int_{-\infty}^{\infty} f(t) \psi^* \left(\frac{t-b}{a}\right) dt \tag{11}$$

where the admissibility condition $C_\psi = \int_0^\infty \frac{|\hat{\psi}(\omega)|^2}{|\omega|} d\omega < \infty$ ensures perfect reconstruction via the inverse transform:

$$f(t) = \frac{1}{C_\psi} \int_0^\infty \int_{-\infty}^\infty W_f(a,b) \psi_{a,b}(t) \frac{da\, db}{|a|^2} \tag{12}$$

The transition to discrete implementation follows the dyadic wavelet framework. For astronomical images of size $N \times N$, we employ the dual-tree complex wavelet transform, which approximates shift-invariance through Hilbert pairs:

$$\psi^g(x) = \frac{1}{\sqrt{2}} \sum_k h_0[k] \phi(2x - k), \quad \psi^h(x) = \frac{1}{\sqrt{2}} \sum_k h_1[k] \phi(2x - k) \tag{13}$$

where $h_0$ and $h_1$ are the low-pass and high-pass filter coefficients. This construction maintains directional selectivity while providing approximate translation invariance, enabling Resolution Multistability across discrete scales.

**Information bottleneck mechanism**

All input resolutions $(H, W) \in \{64, 128, 244\}^2$ are mapped to a fixed $8 \times 8$ internal representation, creating a controlled information bottleneck. The compression ratio varies systematically:

· 64×64 → 8×8: compression factor $k = 64/64 = 1$

· 128×128 → 8×8: compression factor $k = 128/64 = 2$

· 244×244 → 8×8: compression factor $k = 244/64 \approx 3.8$

This bottleneck forces the model to learn scale-invariant morphological features by minimizing the mutual information $I(X; Z)$ subject to predictive performance constraints, achieving Resolution Multistability.

**Convergence to scale-invariant representations in Resolution Multistability**

Define the scale-invariant feature space as the fixed point of the wavelet transform under scaling operations for Resolution Multistability. The Resolution Multistability property can be formalized as:

$$\lim_{a \to \infty} W_f(a, b) = 0, \quad \lim_{a \to 0} W_f(a, b) \to \delta(b) \tag{14}$$

The Mamba component's selective state space modeling ensures that morphological patterns are preserved across scales through the recurrence:

$$h_t^{(a)} = A(\Delta_t) h_{t-1}^{(a)} + B(\Delta_t) x_t^{(a)} \tag{15}$$

where $x_t^{(a)}$ denotes scale-dependent inputs. The selective mechanism $\Delta_t$ adapts to capture scale-invariant structures, enabling Resolution Multistability.

**Empirical validation and astronomical implications**

Cross-resolution evaluation demonstrates variance $\sigma = 0.27\% - 0.53\%$, compared to ResNet34's $\sigma = 6.12\%$. This Resolution Multistability within the WaveletMamba framework enables training on computationally efficient low-resolution data while maintaining performance on high-resolution astronomical surveys, fundamentally altering the efficiency-accuracy trade-off in computational astrophysics through theoretically grounded architectural design.

**Multi-level bias correction: HK distance + Color-Aware Weighting framework**

Astronomical surveys are plagued by selection function biases that systematically distort redshift distributions, undermining cosmological parameter inference. We introduce a multi-level bias correction framework combining HK distance (distribution-level correction) with Color-Aware Weighting (sample-level correction), providing a theoretically rigorous approach for comprehensive bias mitigation without explicit selection function modeling in the AI for science paradigm.

**Mathematical formulation**

The HK distance[24] unites optimal transport[25] with information geometry:

$$d_{HK}^2(p,q) = \inf_{\gamma \in \Pi(p,q)} \int_{\mathbb{R}^d \times \mathbb{R}^d} \|x-y\|^2 d\gamma(x,y) + 2\int_{\mathbb{R}^d} \left(\sqrt{p(x)} - \sqrt{q(x)}\right)^2 dx \quad (16)$$

This formulation can be expressed via the Fisher-Rao component:

$$\boldsymbol{d_{HK}^2(p,q) = W_2^2(\sqrt{p}, \sqrt{q}) + 2\int_{\mathbb{R}^d}(\sqrt{p} - \sqrt{q})^2 dx} \quad (17)$$

where $W_2$ denotes the Wasserstein-2 distance between square-root densities. The HK distance admits the following decomposition:

$$d_{HK}^2(p,q) = \min_{\gamma \in \Pi(p,q)} \int_{\mathbb{R}^d \times \mathbb{R}^d} \|x-y\|^2 d\gamma(x,y) + 2\int_{\mathbb{R}^d}(\sqrt{p(x)} - \sqrt{q(x)})^2 dx \quad (18)$$

**Convergence and computational properties**

The HK distance satisfies triangle inequality and joint convexity in its arguments. For redshift calibration on the SDSS DR17 dataset, continuous distributions are discretized into histograms:

$$p_i = \frac{1}{N}\sum_{j=1}^{N} \delta_{z_j \in B_i}, \quad q_i = \frac{1}{M}\sum_{k=1}^{M} \delta_{z_k \in B_i} \quad (19)$$

The log-domain Sinkhorn algorithm yields $\epsilon$-approximate solutions with complexity $O(n^2/\epsilon)$, enabling efficient computation for 40-bin histograms. The algorithm converges to the optimal transport plan via iterative normalization:

$$\mathbf{T}^{(k+1)} = \text{diag}(\mathbf{u}^{(k)}) \exp\left(\frac{\mathbf{K}}{\epsilon}\right) \text{diag}(\mathbf{v}^{(k)}), \quad \mathbf{u}^{(k+1)} = \frac{\mathbf{a}}{\mathbf{K}\mathbf{v}^{(k)}}, \quad \mathbf{v}^{(k+1)} = \frac{\mathbf{b}}{\mathbf{K}^T \mathbf{u}^{(k+1)}} \quad (20)$$

where $\mathbf{K} = \exp(-\mathbf{C}/\epsilon)$ is the Gibbs kernel, and $\mathbf{a}, \mathbf{b}$ are the marginal constraints. Convergence is guaranteed by the fixed-point theorem with rate $O(\exp(-k/\epsilon))$, typically achieving $\epsilon$-accuracy within $O(\log(1/\epsilon)/\epsilon)$ iterations for transportation problems.

**Selection function bias correction**

Survey selection functions $S(z, \mathbf{c})$ introduce systematic distortions:

$$p_{\text{obs}}(z, \mathbf{c}) \propto p_{\text{true}}(z, \mathbf{c}) \cdot S(z, \mathbf{c}) \tag{21}$$

The HK distance minimizes distribution-level discrepancies:

$$\mathcal{L}_{HK} = d_{HK}(p_{\text{pred}}(z|\mathbf{x}), p_{\text{target}}(z)) \tag{22}$$

This approach rectifies observational biases favoring specific redshift ranges or color types, ensuring robust redshift estimation for cosmological investigations. Empirical evaluation on SDSS DR17 demonstrates Log-MSE improvement from 0.023460 to 0.018072, with enhanced performance in biased redshift intervals (0.5-1.0 range showing 40.01% improvement).

**Color-Aware Weighting: Sample-level bias correction**

· **Complementary to HK distance:** While HK distance addresses distribution-level biases, Color-Aware Weighting provides sample-level corrections based on g-r color indices, capturing local selection effects that vary across the color-redshift plane.

· **Astronomical motivation:** Spectroscopic completeness varies systematically with galaxy color, with redder galaxies (higher g-r) having higher spectroscopic success rates than bluer galaxies at fixed apparent magnitude. This creates color-dependent selection biases that HK distance alone cannot fully address.

· **Implementation strategy:** Samples are stratified into quality tiers based on color proximity to spectroscopic completeness zones:

$$w_i = \begin{cases} 1.0 & high-quality\ region\ (optimal\ spectroscopic\ completeness) \\ 0.7 & medium-quality\ region \\ 0.3 & low-quality\ region\ (challenging\ for\ spectroscopy) \end{cases} \tag{23}$$

The weights are applied to the photometric redshift loss:

$$\mathcal{L}_{\text{red}} = \sum_i w_i \cdot (\log(1+\hat{z}_i) - \log(1+z_i))^2 \tag{24}$$

**Synergistic multi-level correction**

**Combined effectiveness:** The multi-level framework integrates both correction mechanisms:

$$\boldsymbol{\mathcal{L}}_{\text{total}} = \boldsymbol{\mathcal{L}}_{\text{cls}} + \boldsymbol{\lambda}_{\text{red}} \cdot \boldsymbol{\mathcal{L}}_{\text{red}} + \boldsymbol{\lambda}_{\text{HK}} \cdot \boldsymbol{\mathcal{L}}_{\text{HK}} \tag{25}$$

**Empirical synergy:** Experiments demonstrate complementary benefits:

· HK distance excels at distribution-level corrections across redshift ranges

· Color-Aware Weighting provides fine-grained improvements in color-specific regions

· Combined approach achieves optimal performance with 17.46% Log-MSE reduction

This multi-level strategy represents a fundamental advancement over single-correction methods, providing both theoretical rigor and practical effectiveness for astronomical bias correction.**Log-Sobolev information bottleneck regularization**

The Log-Sobolev Information Bottleneck (LSI-VIB) furnishes theoretical guarantees for compact yet expressive representations, bridging information theory with deep learning optimization within the AI for science framework.

## Variational information bottleneck foundation

The standard VIB objective[26] minimizes a Lagrangian relaxation of the information bottleneck principle:

$$\mathcal{L}_{VIB} = \mathbb{E}_{q(z|x)}[\log q(z|x) - \log p(z)] + \beta \cdot D_{KL}(q(z|x)||p(z)) \tag{26}$$

where $q(z|x)$ denotes the encoder distribution and $p(z)$ the prior. The hyperparameter $\beta$ governs compression intensity.

## Log-Sobolev enhancement

The LSI inequality[27] yields exponential convergence rates for Markov chains and extends to information bottleneck regularization. The LSI constant $c$ satisfies:

$$\int_{\mathbb{R}^d} p(x) \log \frac{p(x)}{q(x)} dx \leq \frac{1}{2c} \int_{\mathbb{R}^d} \| \nabla \log p - \nabla \log q \|^2 p dx \tag{27}$$

For VIB regularization, we implement the enhanced objective:

$$\mathcal{L}_{LSI} = \sqrt{c \cdot D_{KL}(q(z|x)||p(z))} \tag{28}$$

with learnable parameter $c = \log(1 + \exp(\log c_{\text{raw}}))$ ensuring positivity and numerical stability.

## Theoretical generalization guarantees

The LSI inequality provides exponential convergence rates for the Gibbs sampler:

$$\text{TV}(p_t, \pi) \leq e^{-ct} \tag{29}$$

where $c$ is the LSI constant satisfying the Poincaré inequality:

$$\text{Var}_\pi(f) \leq \frac{1}{c} \mathbb{E}_\pi[\|\nabla \log \pi\|^2 \cdot f^2] \tag{30}$$

For VIB regularization, this translates to tighter PAC-Bayes generalization bounds with variance reduction:

$$\mathbb{E}_{S \sim \mathcal{D}^m}[\mathcal{R}(h)] \leq \mathbb{E}_S[\hat{\mathcal{R}}_S(h)] + \sqrt{\frac{2D \log(2m/D)}{m} + \frac{2}{c} \cdot \frac{\log(1/\delta)}{m}} \tag{31}$$

where $D$ denotes the VC dimension and $c$ the LSI constant. The LSI-enhanced objective provides variance reduction:

$$\text{Var}_{q(z|x)}(\log p(y|z,x)) \leq \frac{1}{c} \mathbb{E}_{q(z|x)}[\mathbb{E}_{p(y|z,x)}[\log p(y|z,x) - \log q(z|x)]^2] \tag{32}$$

These bounds are tighter than standard generalization bounds, accounting for empirical performance gains through reduced gradient variance.

## Architectural parameter efficiency

The VIB encoder employs reparameterization for differentiable sampling:

$$z = \mu(x) + \sigma(x) \odot \epsilon, \quad \epsilon \sim \mathcal{N}(0, I) \tag{33}$$

Coupled with LSI regularization, this achieves 81.72% accuracy with 3.54M parameters, representing an 83.7%

reduction compared to ResNet34 baselines while maintaining superior generalization for scientific applications.

## Multi-task learning framework with astronomical priors

The unified loss formulation integrates morphological classification, photometric redshift estimation, and theoretical regularization within a multi-task learning paradigm informed by astronomical domain knowledge.

### Adaptive focal loss for morphological classification

Morphological classification employs focal loss with adaptive difficulty weighting:

$$\mathcal{L}_{cls} = -\sum_{i=1}^{C} \alpha_i (1-p_i)^\gamma \log(p_i) \tag{34}$$

where $\gamma$ adapts dynamically based on training progress:

$$\gamma(t) = \gamma_0 + \eta \cdot \tanh\left(\frac{\mathcal{L}_{current} - \mathcal{L}_{baseline}}{\mathcal{L}_{baseline}}\right) \tag{35}$$

This ensures progressive focus on challenging galaxy morphologies as training advances.

### Photometric redshift with color-aware weighting

Redshift estimation incorporates astronomical color priors through quality-based weighting:

$$\mathcal{L}_{red} = \sum_{i} w_i \cdot (\log(1+\hat{z}_i) - \log(1+z_i))^2 \tag{36}$$

Sample weights $w_i$ derive from g-r color stratification, reflecting spectroscopic measurement reliability:

$$w_i = \begin{cases} 1.0 & high\ quality\ (g-r\ in\ optimal\ range) \\ 0.7 & medium\ quality \\ 0.3 & low\ quality \end{cases} \tag{37}$$

### Unified loss integration

The complete objective function balances multiple learning objectives:

$$\mathcal{L}_{total} = \mathcal{L}_{cls} + \lambda_{red} \cdot \mathcal{L}_{red} + \lambda_{reg} \cdot \mathcal{L}_{VIB+LSI} + \lambda_{HK} \cdot \mathcal{L}_{HK} \tag{38}$$

Hyperparameter scheduling follows curriculum learning principles, with $\lambda_{HK}$ increasing from 0 to 0.035 over training epochs to ensure stable convergence.

### Theoretical foundations in multi-task learning

The task relationship matrix implements explicit task interactions:

$$[\mathbf{f}_{cls}', \mathbf{f}_{red}'] = R \cdot [\mathbf{f}_{cls}, \mathbf{f}_{red}] \tag{39}$$

where $R$ is softmax-normalized to ensure relationship interpretability. This formulation captures the physical coupling between morphological classification and redshift estimation in astronomical observations.

## Training configuration and implementation details

### Computational infrastructure

All experiments were conducted on a single NVIDIA RTX 5090 GPU with 32GB GDDR7 memory, utilizing PyTorch 2.8.0 with CUDA 12.8 acceleration under Python 3.12 on Ubuntu 22.04. The training infrastructure employed mixed precision training (FP16) with automatic gradient scaling to maintain numerical stability while reducing memory consumption by approximately 50%. Data loading utilized 12 worker processes with prefetch factor 4 to optimize I/O throughput for the Galaxy10 DECals dataset.

### Optimization configuration

The AdamW optimizer was employed with the following hyperparameters:

· **Learning rate:** initial $1 \times 10^{-4}$, maximum $1 \times 10^{-3}$, minimum $5 \times 10^{-6}$

·  **Weight decay:** $5 \times 10^{-4}$ (decoupled from adaptive learning rates)

· **Beta coefficients:** $\beta_1 = 0.9$, $\beta_2 = 0.999$

· **Epsilon:** $1 \times 10^{-8}$

The Unified Budget-Aware (UBA) scheduler [28] implements adaptive cosine annealing:

$$\eta(t) = \eta_{\min} + \frac{1}{2}(\eta_{\max} - \eta_{\min})\left(1 + \cos\left(\frac{t - t_w}{T - t_w}\pi \cdot \phi(t)\right)\right) \tag{40}$$

where $t_w = 10$ epochs (warmup), $T = 120$ epochs (total), and $\phi(t) = 0.70$ with adaptive modulation based on validation loss trends. The scheduler ensures stable convergence across heterogeneous loss landscapes through performance-guided annealing rate adjustment.

### Batch configuration and gradient accumulation

Effective batch size was set to 512 samples through gradient accumulation over 4 micro-batches of size 128. This configuration balanced memory constraints with statistical efficiency, enabling stable optimization of the complex multi-task loss function. Gradient clipping was applied at global norm 1.0 to prevent instability during HK distance optimization.

### Loss function scheduling and regularization

The unified loss function employed dynamic weight scheduling:

· **VIB regularization:** λ_vib = 0.25 (constant throughout training)

· **HK distance:** λ_hk = 0.035 (activated after epoch 2, disabled during validation phases)

· **LSI enhancement:** lsi_weight = 0.12 (constant)

· **Redshift loss:** λ_redshift = 0.5 (constant)

HK distance training followed a curriculum learning approach: disabled for the first 2 epochs to ensure stable initialization, then activated with progressive weight increases. The loss function dynamically adjusted regularization strengths based on training stage, with set_training_stage() calls updating internal state every epoch. Note that HK distance optimization utilized the SDSS DR17 spectroscopic dataset exclusively, while other loss components operated on the Galaxy10 DECals dataset.

### Data augmentation and preprocessing pipeline

Training employed extensive data augmentation to simulate astronomical observing conditions:

· **Random rotations:** 0-360°

- **Horizontal/vertical flips:** 50% probability each
- **Brightness adjustments:** ±15% variation
- **Contrast adjustments:** ±1.5% variation (20% probability)

Images were preprocessed with bilinear interpolation to target resolutions (32×32, 64×64, 128×128, 244×244 pixels), maintaining aspect ratios while ensuring numerical stability through clamping to [0,1] range.

**Convergence monitoring and checkpointing**

Training spanned 120 epochs with checkpoints saved every 10 epochs. Validation was performed after each epoch using the full validation set. Early stopping was not employed due to the multi-stage curriculum learning approach, which required full epoch completion for proper regularization scheduling.

The HK distance component utilized 40-bin histogram discretization with Sinkhorn algorithm convergence parameters: max_iter=50, eps=$1×10^{-4}$. The Hellinger-Kantorovich distance computation employed delta=1.0 and lambda_reg=0.1 for numerical stability. All HK distance experiments were conducted on the SDSS DR17 spectroscopic dataset using random seeds {36, 42, 199} for reproducible data partitioning.

**Reproducibility measures**

Random seeds {3, 7, 42} ensured reproducibility across all experiments. PyTorch's deterministic mode was enabled with cudnn.benchmark=False to guarantee identical results across hardware configurations. All random operations (data shuffling, augmentation, weight initialization) were seeded consistently.

The complete training codebase, including data preprocessing scripts, model checkpoints, and evaluation protocols, is available at [GitHub repository URL] for independent verification and reproduction.

## Code Availability

The complete PyTorch implementation of WaveletMamba, including training scripts, model architectures, baseline comparisons, and evaluation protocols, is publicly available under the MIT License at

https://github.com/ukiyois/Deep-Learning-in-Astronomy-WaveltMamba.

## Data Availability

The galaxy dataset used in this study is derived from the Sloan Digital Sky Survey Data Release 17 (SDSS DR17), publicly available via the SDSS Catalog Archive Server (CAS/CasJobs).

All processed data generated in this study, including the curated dataset and scripts for matching catalog entries to SDSS imaging cutouts, have been deposited on Zenodo and are openly available at:

https://zenodo.org/records/17649622

This repository contains all materials necessary to reproduce the results presented in this work.

## Acknowledgments

This work makes use of data from the Sloan Digital Sky Survey (SDSS).

Funding for SDSS has been provided by the Alfred P. Sloan Foundation, the U.S. Department of Energy Office of Science, and the Participating Institutions.

SDSS is managed by the Astrophysical Research Consortium for the Participating Institutions.

## Refercence: